\documentclass[letterpaper, 10 pt, conference]{ieeeconf}  % Comment this line out
\IEEEoverridecommandlockouts                              % This command is only
\usepackage{graphics} % for pdf, bitmapped graphics files
\usepackage{epsfig} % for postscript graphics files
\usepackage{url} %%% url 
\usepackage{lastpage} %% pages number
\usepackage{fancyhdr}
\title{\LARGE \bf
Transfer Learning for Prosthetics Using Imitation Learning}
\author{ \parbox{6 in}{ \centering Waleed D. Khamies$^{1}$, Monatser Mohammedalamen$^{1}$, Benjamin Rosman$^{2}$\\
        $^{1}$University of Khartoum, $^{2}$University of Witwatersrand, CSIR\\
        {\tt\small \url{waleed.daud@outlook.com}, \url{montaserfath@gmail.com}, \url{benjros@gmail.com}}}
}
\begin{document}
\maketitle
\thispagestyle{empty}
\pagestyle{empty}
%%%%%%%%%%%%%%%%%%%%%%%%%%%%%%%%%%%%%%%%%%%%%%%%%%%%%%%%%%%%%%%%%%%%%%%%%%%%%%%%
\begin{abstract}
In this paper, We Apply Reinforcement learning (RL) techniques to train a realistic biomechanical model to work with different people and on different walking environments. We benchmarking 3  RL algorithms: Deep Deterministic Policy Gradient (DDPG), Trust Region Policy Optimization (TRPO) and Proximal Policy Optimization (PPO) in OpenSim environment, Also we apply imitation learning to a prosthetics domain to reduce the training time needed to design customized prosthetics. We use DDPG algorithm to train an original expert agent. We then propose a modification to the Dataset Aggregation (DAgger) algorithm to reuse the expert knowledge and train a new target agent to replicate that behaviour in fewer than 5 iterations, compared to the 100 iterations taken by the expert agent which means reducing training time by 95\%. Our modifications to the DAgger algorithm improve the balance between exploiting the expert policy and exploring the environment. We show empirically that these improve convergence time of the target agent, particularly when there is some degree of variation between expert and naive agent.
\end{abstract}
%%%%%%%%%%%%%%%%%%%%%%%%%%%%%%%%%%%%%%%%%%%%%%%%%%%%%%%%%%%%%%%%%%%%%%%%%%%%%%%%
\section{INTRODUCTION}
\subsection{ Problem Definition}
Limbs are hugely valuable to many people, in that they improve mobility and the ability to manage daily activities, as well as provide the means to stay independent. It is costly (50K USD) and time-consuming for the manufacturers to design artificial limbs customized for one person, Designing intelligence prosthetics which deal with the large differences between humans (like human body dimensions, weights, height and walking styles) is so complicated by the large variability in response among many individuals. One key reason for this is that our understanding of the interactions between humans and prostheses is not well-understood, which limits our ability to predict how a human will adapt his or her movement. Physics-based, biomechanical simulations are well-positioned to advance this field as it allows for many experiments to be run at low cost.
\subsection{Environment}
We use OpenSim “ProstheticsEnv” environment, which models one human leg and prosthetic in another leg see in fig(1), OpenSim is a 3D human model simulator, which consists of observations of joints, muscles and tendons, 19 actions, and the reward $R_{t}$ is the negative distance from the desired velocity in eq(~\ref{eq:reward}).
\begin{figure}[ht!]
\begin{center}
\includegraphics[width=0.3\textwidth]{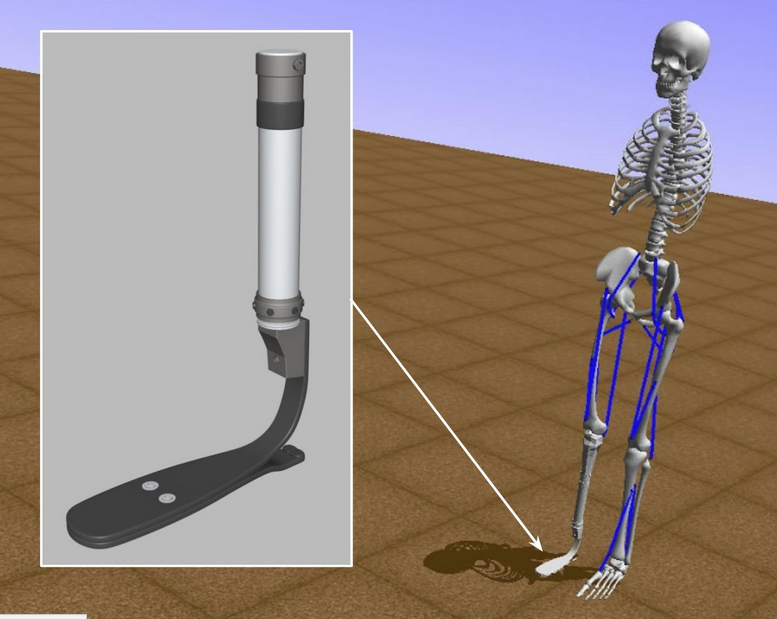}
\label{fig:OpenSim ProstheticEnv Environment}
\caption {OpenSim ProstheticEnv Environment.}
\end{center}
\end{figure}
\begin{equation}
{R_{t}=9-(3-V_{t})^2}
\label{eq:reward}
\end{equation}
where $V_{t}$ is the horizontal velocity vector of the pelvis.
OpenSim environment has a limitation it is very slow to run due to the high number of observations and state variables.

\subsection{Reinforcement Learning (RL) algorithms}
Reinforcement Learning (RL) will help prosthetics to calibrate with differences between humans and differences between walking environments \cite{c1}, RL is a machine learning paradigm, where an agent learns the optimal policy for performing a sequential decision making without complete knowledge of the environment \cite{c2}. The agent must explore the environment by taking action $A_{t}$ and edit the policy according to the reward function to maximize the reward $R_{t}$.
We use the DDPG algorithm \cite{c3}, TRPO and PPO to train the agent. DDPG is a model-free, off-policy actor-critic algorithm using deep function approximators that can learn policies in high-dimensional, continuous action spaces.
\subsection{Imitation Learning}
The main problem with RL algorithms is the time needed to solve the problem -training time- because the algorithm must explore the environment and adapt its policy according to the reward at every timestep. Imitation learning is a specific subset of RL where the learner tries to mimic an expert’s action in order to achieve the best performance. The main advantage of DAgger is that the expert teaches the learner how to recover from past mistakes \cite{c4}, and we aim to leverage this to illustrate behaviour learning. 
There are many ways to accelerate the learning process in RL, such as Cross-Domain Transfer \cite{c2}, Inter-task Mapping via Artificial Neural Network (ANN) \cite{c5}. We use Imitation learning to achieve that by implementing DAgger algorithm. The DAgger algorithm has shown to be able to achieve expert-level performance after a few data aggregation iterations \cite{c6}.
To use imitation learning there are two assumptions:
\begin{enumerate}
    \item Similarity between the expert and the target agent in actions, observations space and the reward function.
    \item Environment must be described by a Markov Decision Process (MDP).
\end{enumerate}
In the standard DAgger algorithm, the target agent exploits the expert policy and stops exploring the environment. This may be a problem, as the target agent should balance between exploitation and exploration. We propose some improvements to the DAgger algorithm to encourage the exploration.
\section{Experiments}
We run the following experiments:\footnote{codes available at \url{github.com/montaserFath/Reinforcement-Learning-for-Prosthetics}}
\begin{enumerate}
    \item Run RL algorithms (DDPG, TRPO and PPO) in OpenSim ProstheticsEnv, 2,000 episodes and 1,000 timesteps in the episode to give an agent more time to walk or stand up.
    \item we trained a DDPG agent to achieve positive reward (around +100) in the standing up task.
    \item we use that agent as an expert to evaluate the DAgger algorithm.
    \item we modify DAgger so that the expert agent labels the target agent’s actions based on the timestep reward, by comparing between the timestep reward of the expert agent and the target agent on a given timestep: if the expert agent has less reward than the target agent, the expert keeps the target agent’s action and the opposite is true.
    \item we use the target action value instead of timestep reward to do the comparison, and we sum the timestep rewards from a given state and action pair until the end of the episode
    \item we used the epsilon-greedy method \cite{c2}, where the algorithm has the choice to select between taking the target action with a probability $ 1-\epsilon $ or the expert action with a probability $\epsilon$.
\end{enumerate} 
\section{Results}
\begin{figure}[ht!]
\begin{center}
\includegraphics[width=0.4\textwidth]{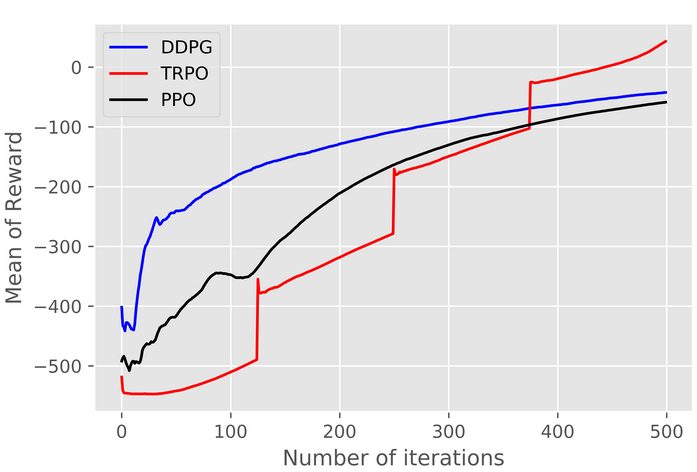}
\caption{Number of Iterations VS Reward Mean in OpenSim ProstheticsEnv Environment.}
\label{fig:Number_of_Iterations}
\end{center}
\end{figure}

\begin{figure}[ht!]
\begin{center}
\includegraphics[width=0.45\textwidth]{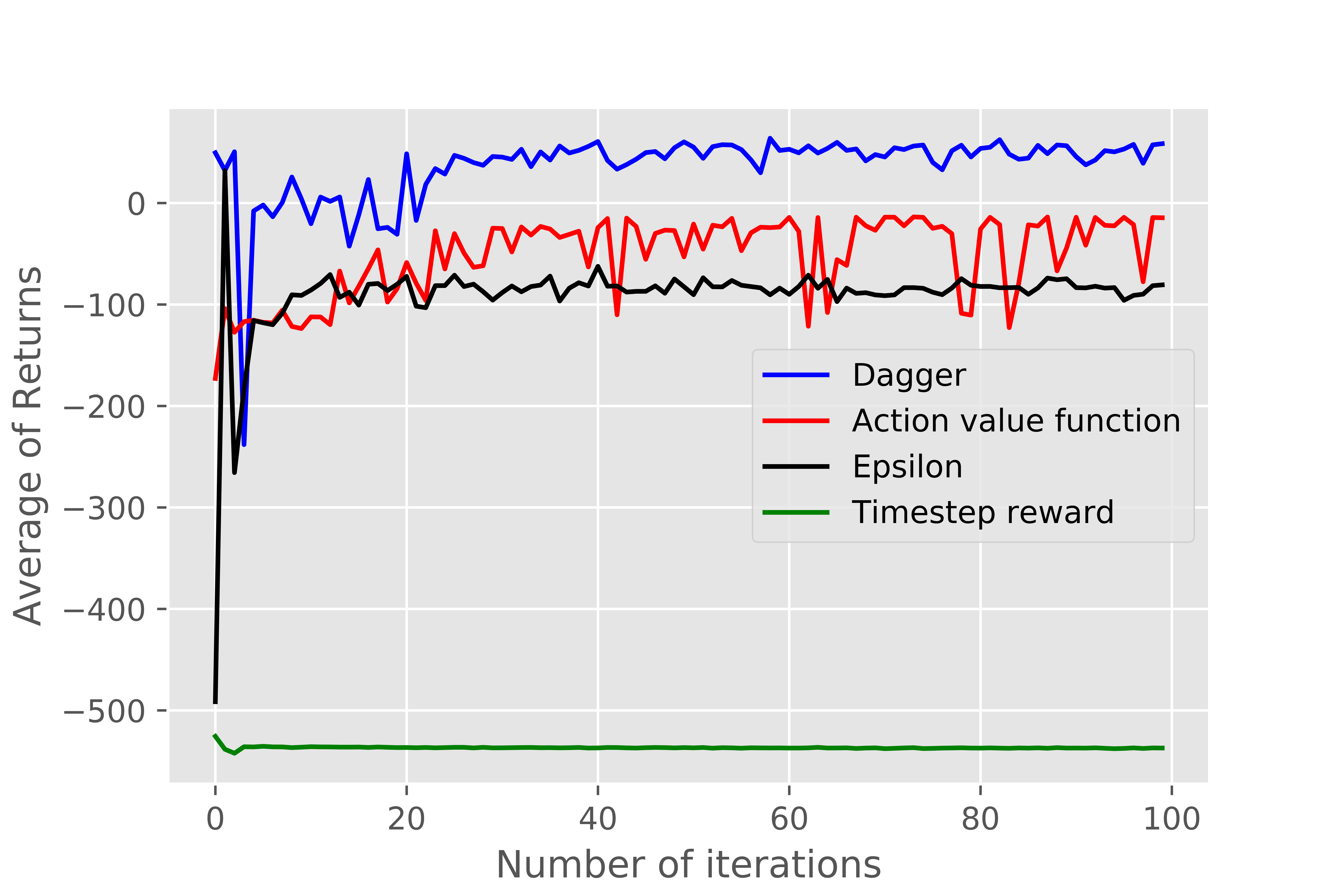}
\caption{Comparison between DAagger algorithm, Timestep Reward, action value and Epsilon-greedy}
\label{fig:imitation_learning}
\end{center}
\end{figure}

The maximum reward mean achieved by TRPO (see table~\ref{table:OpenSim_ProstheticsEnv} and fig~\ref{fig:Number_of_Iterations}), but it takes more time comparing with PPO and DDPG because it need to find the inverse of matrix which takes time. Although of this reward the agent can not walk for more than one step and sometimes it falls before the first step.
Dagger algorithm achieved the best average reward comparing with other algorithms which balance between exploiting and exploring (see fig~\ref{fig:imitation_learning}), we think the reasons behind that:
\begin{enumerate}
    \item The expert policy has a high reward.
    \item The high similarity between expert and naive agent.
    \item The naive agent needs more time to run by increasing the number of iterations.
\end{enumerate} 
The main problem with timestep reward modification, it compares timestep reward (short-term) adding to this the large variation between timesteps. When the variation between episodes is small in Action-value.
The naive agent has gotten a reward greater than the expert agent, so roles can be exchanged, the expert can be naive and the naive can be an expert which will decrease training time significantly.
%%%%%%%%%%%%%%%%%%%%%%table%%%%%%%%%%%%%%%%%%%%%%%%
\begin{table}[ht!]
\begin{center}
\begin{tabular}{|c|c|c|}
 \hline
\textbf{Algorithm} & \textbf{Maximum reward} & \textbf{Mean reward} \\  \hline
\textbf{DDPG} &  113 & -42 \\  \hline
\textbf{TRPO} & 194 & 43 \\  \hline
\textbf{PPO} & 70 & -58 \\  \hline
\end{tabular}
\caption{Comparison between algorithms in ProstheticsEnv.}
\label{table:OpenSim_ProstheticsEnv}
\end{center}
\end{table}
%%%%%%%%%%%%%%%%%%%%%%%%%%%%%%%%%%%%%%%%
%%%%%%%%%%%%%%%%%%table%%%%%%%%%%%%%%%%%%%%%%
\section{CONCLUSIONS}
We have applied imitation learning in a humanoid environment to accelerate the learning process. The naive agent reaches convergence within 5  iterations while the expert reaches it after 100 iterations which means reducing training time by 95\%. 
The DAgger algorithm achieved the best average reward comparing with other algorithms which balance between exploiting and exploring, these algorithms will work better when there is some degree of variation between expert and naive agent and this is what we are planning to do in the future by apply imitation learning from normal human legs to prosthetic, the main challenge will be how to figure out the differences and similarities between it.
\section{Research Limitations}
\begin{enumerate}
    \item The prosthetic model can not walk for large distances even can falls before completing the first step.
    \item Each experiment runs for one time, So we are planing to repeat each experiment number of times with different random seeds and take the average and variance.
    \item We used same hyperparameters for all algorithm to benchmark algorithms, we need to select the best hyperparameters for each algorithm and environment.
    \item We benchmarcked three RL algorithms only and from one library(ChainerRL). So we are planing to use different implementations.
\end{enumerate}
\addtolength{\textheight}{-12cm}   % This command serves to balance the column lengths

\end{document}